\pdfoutput=1

\documentclass[11pt]{article}

\usepackage[final]{acl}

\usepackage{times}
\usepackage{latexsym}

\usepackage[T1]{fontenc}

\usepackage[utf8]{inputenc}

\usepackage{microtype}

\usepackage{inconsolata}

\usepackage{subcaption}

\usepackage{caption}
\usepackage{graphicx}
\usepackage{float}
\usepackage{svg}
\usepackage{amsmath} 
\usepackage{lipsum}
\usepackage{colortbl}
\usepackage{booktabs}
\usepackage{nicematrix}
\usepackage[listings]{tcolorbox}
\usepackage[a4paper]{geometry}
\usepackage{changepage}

\definecolor{mycustomcolor}{HTML}{1F77B4}

\title{Critical Questions Generation: \\ Motivation and Challenges}

\author{Blanca Calvo Figueras \\
  HiTZ Center - Ixa \\ University of the Basque \\ Country UPV/EHU \\
  \texttt{blanca.calvo@ehu.eus} \\\And
  Rodrigo Agerri \\
  HiTZ Center - Ixa \\ University of the Basque \\ Country UPV/EHU \\
  \texttt{rodrigo.agerri@ehu.eus} \\}

\begin{document}
\maketitle
\begin{abstract}

The development of Large Language Models (LLMs) has brought impressive performances on mitigation strategies against misinformation, such as counterargument generation. However, LLMs are still seriously hindered by outdated knowledge and by their tendency to generate hallucinated content. In order to circumvent these issues, we propose a new task, namely, \textit{Critical Questions Generation}, consisting of processing an argumentative text to generate the critical questions (CQs) raised by it.
In argumentation theory CQs are tools designed to lay bare the blind spots of an argument by pointing at the information it could be missing.
Thus, instead of trying to deploy LLMs to produce knowledgeable and relevant counterarguments, we use them to question arguments, without requiring any external knowledge.
Research on CQs Generation using LLMs requires a reference dataset for large scale experimentation. Thus, in this work we investigate two complementary methods to create such a resource: (i) instantiating CQs templates as defined by Walton's argumentation theory and (ii), using LLMs as CQs generators. By doing so, we contribute with a procedure to establish what is a valid CQ and conclude that, while LLMs are reasonable CQ generators, they still have a wide margin for improvement in this task.

\end{abstract}

\section{Introduction}

Natural Language Processing (NLP) applications to deal with misinformation have become a popular line of research in tasks such as fact verification \cite{thorne_fever_2018}, evidence retrieval \cite{soleimani_bert_2019} or counterargument generation \cite{chung_conan_2019,chen2023exploring}.
However, even when deploying generative Large Language Models (LLMs), most applications face challenges regarding three issues: LLMs often lack the required up-to-date knowledge for these tasks \cite{gao2023retrieval}, there is not always an agreement on what is the truth \cite{chang_detecting_2024}, and LLMs themselves can produce hallucinations or rely on unfaithful data, generating misinformation of their own making \cite{xu_hallucination_2024,lin_truthfulqa_2022}.

Yet, instead of requiring the LLMs to output factual knowledge, could we use them to point at the missing or potentially uninformed claims? In other words, could we use LLMs to uncover the blind spots in the argumentation? To open this line of research, we ground our work on argumentation theory, which has for centuries been studying dialogical exchanges of information. Specifically, we look into \textit{argumentation schemes}, a set of abstract structures developed by systematically identifying common patterns of argumentation and outlining the defeasibility of these patterns. In these structures, the devices designed to find the blind spots in the arguments are called \textit{critical questions}. 

\begin{figure*}[h!]
\footnotesize  
\noindent\rule[0.5ex]{\linewidth}{1pt}
\begin{subfigure}{.49\linewidth}
\caption{\textbf{Scheme -- Argument from Cause to Effect}}
\textbf{Premise:} Generally, if people pour into the USA, then Americans lose their jobs. \\
\textbf{Premise:} In the current situation, people are pouring into the USA.\\
\textbf{Conclusion:} In the current situation, Americans lose their jobs.  \\ \\ \\
\textbf{CQ:} How strong is the generalisation that if people pour into the USA then Americans will lose their jobs?  \\
\textbf{CQ:} Are there other factors in this particular case that could be interfering with the fact that Americans lose their jobs? \\ \\ 

\end{subfigure}\hfill
\begin{subfigure}{.49\linewidth}
\caption{\textbf{Scheme -- Practical Reasoning}}
\textbf{Premise:} There is the goal of making the economy fairer. \\
\textbf{Premise:} Raising the national minimum wage is a means to realize the goal of making the economy fairer.\\
\textbf{Conclusion:} Therefore, raising the national minimum wage ought to occur.  \\ \\

\textbf{CQ:} Are there other relevant goals that conflict with making the economy fairer?  \\
\textbf{CQ:} Are there alternative actions to raising the national minimum wage to achieve making the economy fairer? If so, which is the most efficient action?  \\
\textbf{CQ:} Could raising the national minimum wage have consequences that we should take into account? Is it practically possible?
\end{subfigure}
\noindent\rule[0.5ex]{\linewidth}{1pt}
\caption{Arguments from the US2016 dataset \cite{visser_annotating_2021}, instantiated using the templates of argumentation schemes and critical questions defined in \citet{walton_argumentation_2008}.}
\label{example}
\end{figure*}

Critical questions are the set of inquiries that could be asked in order to judge if an argument is acceptable or fallacious. Therefore, these questions are designed to unmask the assumptions held by the premises of the argument and attack its inference. In the theoretical framework developed by \citet{walton_argumentation_2008}, argumentation schemes are represented as templates depicting the premises, the conclusion, and the critical questions of each scheme. This framework is useful to promote critical thinking, since it allows uncovering fallacies by answering questions. Figure \ref{example} shows two examples of argumentation schemes and their corresponding critical questions (CQs). The first of these examples is an argument that links a cause (migration) to an effect (unemployment). Therefore, the CQs related to this argument ask about the strength of this relation and the possibility of other causes also having a role in the effect. The second example fits the scheme of \textit{practical reasoning}. That is, given a goal, the argument defines an action to achieve it. Here, the CQs ask about the compatibility of this goal with others, the alternative actions to achieving this goal, and the potential consequences of the proposed action. 

Previous work has proved the usefulness of CQs for enhancing fallacy identification \cite{musi_developing_2022}, and for argumentative essays evaluation \cite{song_applying_2014}. But, to the extent of our knowledge, there has not been any attempt to automate the generation of CQs. In this work, we propose the task of \textit{Critical Questions Generation}: given an argumentative text, the model is asked to generate the necessary CQs to assess the acceptability of the arguments in the text. In this setting, the argumentative text is the input and the set of CQs is the target output. As in other NLP tasks, such as machine translation or paraphrasing, the model is not required to find new information, but to understand and reformulate the input in a certain way.

A crucial requirement to investigate the automatic generation of CQs is to have reference data for experimentation. However, as far as we know, there  has not been any attempt to create such a resource. In order to address this shortcoming, in this paper we investigate two methods for creating a dataset for the generation of CQs: (1) using the sets of CQ templates defined in \citet{walton_argumentation_2008}'s theory  (from now on, theory-CQs); and (2) using LLMs to generate these CQs (from now on, llm-CQs). While looking into these methods, we attempt to answer the following research questions: (i) are current Large Language Models good critical question generators? (ii) how can we operationalize what is a valid critical question? (iii) what is the optimal strategy to build a reference dataset for large scale experimentation on the task of \textit{Critical Questions Generation}?

To answer these questions, we start by looking at the theoretical sets of CQs and instantiating them using a set of argumentative texts already annotated with argumentation schemes \cite{visser_annotating_2021, lawrence_bbc_2018}. As a second step, we prompt two state-of-the-art LLMs to give us candidate CQs for these same argumentative texts, and we design a procedure to evaluate their relevance towards the texts and their validity as CQs. We then compare the two methods and highlight the main challenges faced by LLMs when generating CQs. Summarizing, the main contributions of this work are: 

\begin{itemize}
    \item We propose the task of \textit{Critical Questions Generation} and motivate it by relying on previous work.
    \item We use naturally-occurring dialogical data to study how to generate critical questions using the theory templates and LLMs.
    \item We operationalize how to define a valid critical question.
    \item We study the main challenges faced by LLMs when generating critical questions.
\end{itemize}

In this work, we observe that questions generated using theory and questions generated using LLMs are complementary: while theory-CQs are mostly about relations between premises, llm-CQs rather ask about evidences. Additionally, LLMs introduce a new type of questions: those asking about further definition of the terms used in the arguments. Regarding the performance of current LLMs, we observe that models struggle to output relevant CQs and output many non-critical questions. Therefore, we conclude that more advanced training and prompting techniques should be used and, to this end, reference data should be created using both the theory and LLMs' methods. All the data and code in this project has been released.\footnote{\url{https://github.com/hitz-zentroa/critical\_questions\_generation}}

\section{Previous Work}\label{sec:motivation}

To contextualise this work, we discuss the relation between argumentation and misinformation, introduce the nature of critical questions, and offer related work on argumentation schemes from a computational point of view.

\subsection{Using argumentation to fight misinformation}\label{sec:misinfo}

Misinformation has been tackled using many strategies: from debunking strategies (e.g. fact-checking propagated information) to pre-bunking (e.g. exposing disinformation strategies to make citizens resilient towards manipulation). However, recent studies have shown that pre-bunking has a potentially longer effect, since the learned skills are not bound to specific contexts \citep{maertens_long-term_2021}. Following this, digital applications have been built to enhance citizens' abilities to deal with misinformation, such as the recognition of misleading sources and headlines (Fakey,\footnote{\url{https://fakey.osome.iu.edu/}} NewsWise headlines quizz\footnote{\url{https://www.theguardian.com/newswise/2021/feb/04/fake-or-real-headlines-quiz-newswise-2021}}), the identification of fake images (Real or Photoshop quizz\footnote{\url{https://landing.adobe.com/en/na/products/creative-cloud/69308-real-or-photoshop/}}), or the decision-making processes of news rooms (BBCireporter,\footnote{\url{https://www.bbc.co.uk/news/resources/idt-8760dd58-84f9-4c98-ade2-590562670096}} NewsFeed Defenders\footnote{\url{https://www.icivics.org/games/newsfeed-defenders}}). 

However, these applications focus mostly on dealing with fake information, while misinformation is often generated by drawing invalid relations between claims and the premises provided to support these claims \citep{musi_developing_2023}. In this sense, more recent pre-bunking applications have focused on techniques based on argumentation theory, which have the goal of evaluating the connections between the available evidence and the statement that it is trying to support \citep{lawrence_bbc_2018, visser_argumentation_2020, de_liddo_lets_2021, altay_scaling_2022}.

In this line of research, \citet{musi_developing_2023} developed a chatbot that, following gamification principles, used a dialogical context to teach users how to identify fallacies by being exposed to critical questions. Users of this tool showed an overall increased ability to identify fallacious arguments. 
While the scenarios portrayed in Musi's chatbot are based on an annotated database of 1,500 fact-checked news, 
latest NLP advances in LLMs could be used to generate critical questions on unseen arguments, therefore being able to use this tool to deal with any upcoming domain.

Applications of language models in the fight against misinformation have often been framed as classification and information retrieval tasks \cite{montoro_montarroso_fighting_2023}. In contrast, we propose to use LLMs as a tool for generating questions, which enhances the relativistic conceptions of truth of most critical thinking paradigms \citep{musi_developing_2023}, as opposed to the absolutist notions of truth encouraged by using LLMs as question-answerers and classifiers. 

\subsection{The nature of critical questions}\label{sec:critical_questions}

Critical questions are an essential element of the notion of \textit{argumentation schemes}. Argumentation schemes are ``forms of arguments (structures of inference) that represent structures of common types of arguments used on everyday discourse'' \cite{walton_argumentation_2008}. These arguments are defeasible, meaning that their conclusions can be accepted only provisionally while there is no evidence that defeats it. Defeasible arguments are the most common arguments in everyday discussions, and knowing what to ask before accepting them is an important skill.

The predecessor of argumentation schemes were topics (\textit{topoi} in Aristotle's Rhetoric), which were conceived as warrants that back the logical inferences drawn from premises to conclusions. Modern researchers have adapted them for use in computational applications \cite{reed_applications_2001, macagno_argumentation_2017}. Additionally, these tools have become popular among critical thinking researchers for their pedagogic usefulness.

In pedagogical terms, argumentation schemes can be used ``as a way of providing students with additional structure and analytic tools with which to analyze natural arguments and to evaluate them critically'' \cite{walton_argumentation_2008}. In this approach, critical questions function as memory devices: a way to recall the missing information in the argument.

Although the goals and usefulness of critical questions have been extensively discussed, up to our knowledge, there has not been any successful attempt to operationalize what is and what is not a valid critical question. Since our goal is to create them automatically, setting this boundary becomes a necessary first step.

Most definitions of critical questions are highly linked to their function. Following this tradition, it could be argued that a good critical question is the one that fulfills its goal: pointing at reasons to \textit{rebut the argument}. Moreover, critical questions can not only attack the acceptability of an argument by defeating its conclusion, but also undercut it by attacking the connection between the premises and the given conclusion \cite{pollock_defeasible_1987}. In Section \ref{sec:methodology}, we operationalize this definition of valid CQ, and in Section \ref{sec:results}, we implement it in the evaluation of llm-CQs.

\subsection{Argumentation Schemes in Computational Argumentation}
\label{sec:computational_args}

While no attempt exists to automatically generate CQs, there has been some work on argumentation schemes annotation and detection, which we will be taking as a starting point.

One of the most ambitious works in argumentation from a computational point of view was the Araucaria project, which created a database of arguments annotated in Argument Markup Language that included argumentation schemes \cite{reed_language_2008}. Later, the Inference Anchoring Theory (IAT \citet{budzynska2011whence}) became a popular format for representing how arguments are created in dialogical settings. IAT diagrams feature locutions, propositions, dialogical relations, and propositional relations. Recent work has also added argumentation-scheme labels to IAT diagrams. The available datasets annotated with IAT and schemes are listed in Table \ref{tab:data}. 

\begin{table*}[]
 \footnotesize
\begin{tabular}{lccccc}
\hline
\bf Name & \bf Paper &  \bf Nº Args. & \bf Nº Schemes & \bf Original Format & \bf Domain  \\
\hline
US2016 & \citet{visser_annotating_2021} & 413 & 60 & Oral debate & Politics \\
Moral Maze & \citet{lawrence_bbc_2018} & 79 & 32 & Oral debate & Politics \\
US2016reddit &  & 19 & 4 & Written social media & Politics \\
EO\_PC & \citet{lawrence_combining_2015} & 139 & 3 & Written & Not specified \\
Reg. Room Div. & \citet{konat_corpus_2016} & 227 & 7 & Written social media & Product Regulations  \\
Legal &  & 545 & 12 & Written & Legal \\
\hline
\end{tabular}
\caption{Available data in IAT format with argumentation schemes. All the datasets are in English.}
\label{tab:data}
\end{table*}

Other datasets that are labeled with argumentation schemes although not in the IAT format are the social media datasets from \citet{jo_classifying_2021}, which contain 1,924 examples of 2 argumentation schemes; and the Genetics Research Corpus, which identifies argumentation schemes in scientific claims from genetic research articles \cite{green_identifying_2015}. Lately, datasets with synthetic arguments have been released \cite{kondo_bayesian_2021, ruiz-dolz_nlas-multi_2024, saha_argu_2023}. However, we are interested in naturally-occurring arguments. 

The task of automatically identifying argumentation schemes was first attempted by \citet{feng_classifying_2011} and \citet{lawrence_argument_2016}, using machine learning techniques. Later, \citet{jo_classifying_2021} used logic and theory-informed mechanisms for a similar task, and \citet{kondo_bayesian_2021} used language models, showing the difficulty of identifying schemes (with 7 categories, their overall accuracy with BERT \cite{devlin_bert_2019} was  27.5\%). 

In previous work, it has been observed that tasks requiring complex reasoning remain a challenge for LLMs \cite{xu_are_2023, gendron_large_2024, han_folio_2022}. Furthermore, \citet{payandeh_how_2023} demonstrated that LLMs are easily convinced using logical fallacies, and \citet{ruiz-dolz_detecting_2023} showed that LLMs fail when asked to detect argumentative fallacies. The task of fallacy detection is highly related to our work \cite{sahai_breaking_2021, goffredo_fallacious_2022, alhindi_multitask_2022, helwe-etal-2024-mafalda}. However, in this work we wish to foster human-computer interaction and use LLMs to raise the questions that would help a human unmask the fallacies of its caller.

So far, the most similar work to ours is \citet{musi_developing_2023}, where they developed a chatbot that outputted critical questions from a database of possible issues, and \citet{song_applying_2014}, where they found that human annotations identifying the CQs present in essay evaluations contributed significantly to predicting the grade. While their experiments tested the usefulness of using CQs, none of these two tried to generate them automatically. 

\section{Data}
\label{sec:data}

For the purpose of this work we have decided to use a subset of the US2016 \cite{visser_annotating_2021} and the Moral Maze datasets \cite{lawrence_bbc_2018}, which, as explained in the previous section, have already been transcribed and annotated with argumentation schemes. Both of these datasets are oral debates, and they are structured as sequences of interventions by different debaters. 

In order to use these datasets, we have mapped their labels to the argumentation schemes in \citet{walton_argumentation_2008}. Since the labels of both of these datasets are based on Walton's work, the mapping has amounted to terminology matching. Given the long list of argumentation schemes, we have decided to work with the 18 most frequent schemes. Annex \ref{annex:distribution} provides the mapping and the distribution of argumentation schemes for each of the two datasets.

Since both of these datasets have been annotated as IAT diagrams, each argumentation scheme label links two or more propositions in the debate, forming an argument.\footnote{For a comprehensive explanation of IAT diagrams see \citet{hautli-janisz_qt30_2022}.} The debates are composed of interventions, which we are going to use as our \textit{argumentative texts}. Each intervention can have many annotated arguments (or none). 
After preprocessing,\footnote{We structure the data by intervention, splitting the very long interventions, and merging the very short ones (for an example, see the columns "Intervention" in Table \ref{tab:annotation} and Figure \ref{fig:main}). The code on how to go from the IAT diagrams to our dataset has been published on Github.}
we obtain 370 interventions (73 from Moral Maze and 297 from US2016) of which 117 contain at least one argument (25 from Moral Maze and 92 from US2016).

 For the manual analysis of this work, we use 21 of the interventions, chosen to keep the label distribution as similar as possible to the one in the full datasets; 10 of these interventions come from US2016 and 11 from Moral Maze. The distribution of the 60 arguments contained in these 21 interventions can be found in Annex \ref{annex:distribution_sample}. 

\section{Our Method}\label{sec:methodology}

In order to identify the challenges in the task of \textit{Critical Questions Generation}, there is an urgent need for reference data. To explore how this data should be created, we generate critical questions both using the theory templates and LLMs.

\begin{figure*}[h]
    \centering
    \includegraphics[width=0.7\textwidth]{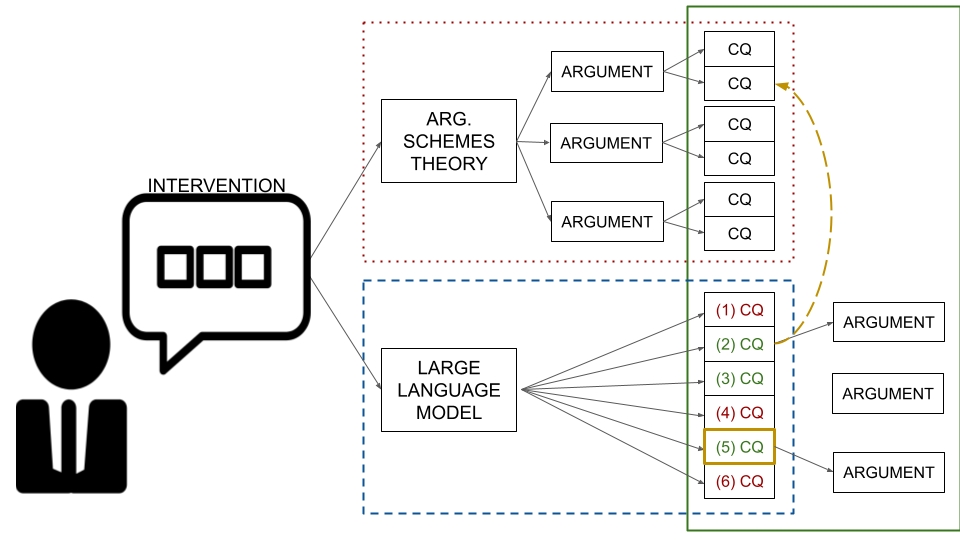}
    \caption{Outline of the steps taken in our approach. Starting from each intervention, we generate CQs using the theory templates (red-dotted box) and the LLMs (blue-dashed box). In the green box, we relate the relevant llm-CQs to the arguments of the intervention (if possible), and relate these llm-CQs to a theory-CQ (if possible). }
    \label{fig:methodology}
\end{figure*}

To generate CQs based on Walton's theory (theory-CQs), we take each annotated argument and instantiate the CQs associated to that argumentation scheme (red-dotted box at the top of Figure \ref{fig:methodology}). Regarding the generation of CQs with LLMs (llm-CQs), we prompt two state-of-the-art LLMs and we evaluate the relevance of the candidate CQs towards the argumentative text (blue-dashed box at the bottom of Figure \ref{fig:methodology}). We then relate the llm-CQs to the arguments of the text and to the theory-CQs (green box), and assess the validity of the llm-CQs that relate to an argument but do not correspond to any of the existing theory-CQs (such as CQ 5 in Figure \ref{fig:methodology}). 
In the rest of the section, we describe in detail each of these processes.

\subsection{Generation using theory}
\label{sec:theory}

The critical questions based on theory are defined using the set of CQs in \citet{walton_argumentation_2008}. We reformulate some of these questions to make them sound more natural (the final set can be found in Annex \ref{annex:questions}). To transform these questions into tailored CQs for each argument, we first manually annotate the text needed to fill the gaps of the variables in the argumentation-schemes' templates. For each argument, the annotator sees the premises and conclusion associated with the argumentation scheme (i.e. the template), the propositions of the argument, and the entire intervention in which the argument occurred. For instance, to annotate argument \textit{a} in Figure \ref{example}, the annotator saw the data in Table \ref{tab:annotation}, and was asked to write the text that is needed to instantiate the scheme template. In this case, $<eventA>$ = "people are pouring into the USA" and $<eventB>$ = "Americans might lose their jobs". 

\begin{table*}[h]
 \scriptsize \centering
\begin{tabular}{c|c|c|c}
\hline
\bf \begin{tabular}[c]{@{}c@{}}\textbf{Argument}\\ \textbf{Scheme}\end{tabular} & \bf Scheme Template &  \bf Propositions & \bf Intervention  \\
\hline
\begin{tabular}[c]{@{}c@{}} Argument\\ from \\ CauseToEffect \end{tabular}
& 
\begin{tabular}[l]{@{}l@{}} Generally, if $<eventA>$, then $<eventB>$.  \\ In the current situation, $<eventA>$. \\  In the current situation, $<eventB>$. \end{tabular}
& 
\begin{tabular}[c]{@{}c@{}} "people are pouring \\ into the USA" \\ \& \\"Americans are losing \\ their jobs" \end{tabular}
 &
 \begin{tabular}[c]{@{}c@{}} 
TRUMP: I want to make America great again \\
We are a nation that is seriously troubled \\
We're losing our jobs \\
People are pouring into our country \\
The other day , we were deporting 800 people \\
perhaps they passed the wrong button \\
they pressed the wrong button \\
perhaps worse than that \\ 
it was corruption [...] \end{tabular}
\\
\hline
\end{tabular}
\caption{Data seen by the annotator when defining the variables to fil the argumentation scheme templates. Propositions and argumentation schemes come from the IAT annotations in the US2016 dataset \cite{visser_annotating_2021}. The scheme template comes from \citet{walton_argumentation_2008}.}
\label{tab:annotation}
\end{table*}

We used two annotators for this task, and achieved an inter-annotator agreement (IAA) of 0.88 with a sample of 174 variables.\footnote{The extended explanation of this annotation will be published as guidelines.} In the end, 9 arguments were discarded by both annotators, as they were not able to find the connection between the propositions and the argumentation scheme that had been given to its relation.

We then instantiated the CQs, substituting each variable for the piece of text that had been annotated. This step resulted in questions with grammatical errors, which we post-edited manually, with 39.44\% of the questions getting editions. We discarded 10 of the questions for being meaningless. Most common corrections consisted of modifying verbs from infinitive to gerund forms and vice-versa, or from singular to plural forms and vice-versa, and removing double negations. This process resulted in the generation of 129 theory-CQs associated to 51 arguments, an average of 6.14 CQs per intervention.

\subsection{Generation using LLMs}\label{sec:llm-generation}

Walton's sets of critical questions are thought of as starting points towards rebuttal strategies. However, they do not intend to be an exhaustive list of the potentially useful CQs for each scheme \cite{walton_nature_2005}.

For this reason, it is interesting to experiment with LLMs to see if the models can generate questions that are valid CQs but are not included in Walton's templates. In this sense, our goal is to have a list of valid CQs as exhaustive as possible that could be used as reference data. However, llm-CQs should be carefully curated. For this purpose, we have designed a method to filter the candidate llm-CQs and obtain a list of relevant and valid CQs. This procedure will serve, at the same time, as an evaluation of how good current LLMs are at generating CQs.

To this goal, we prompt two LLMs to generate the CQs that each intervention may arise in a zero-shot setting. We experiment with two different prompts, one including the query and the intervention,\footnote{Prompt 1: \textit{List the critical questions that should be asked regarding the arguments in the following paragraph: $<SPEAKER>$: “$<INTERVENTION>$”}} and one that also includes a definition of critical questions.\footnote{Prompt 2: \textit{Critical questions are the set of enquiries that should be asked in order to judge if an argument is good or fallacious by unmasking the assumptions held by the premises of the argument. List the critical questions that should be asked regarding the arguments in the following paragraph: $<SPEAKER>$: “$<INTERVENTION>$”}} Then, the evaluation process to filter the candidate llm-CQs has the following three steps.

First, we manually review each of the candidate CQs to detect those that are not relevant with respect to the given argumentative text (i.e. the intervention). We have detected three issues that make the questions not relevant: (a) \textbf{the introduction of new concepts or topics} -- ideally LLMs should generate CQs related to the content of the intervention, not introducing new topics or concepts that may carry the model's biases; 
(b) \textbf{bad reasoning}, namely, questions critical towards positions or claims the speaker does not hold; or (c) \textbf{non-specific critical questions} that could be asked on any argument and that do not take the intervention into account.

Second, using the set of relevant llm-CQs, we match each of these to one of the annotated arguments (if possible), and then assess if the matched CQs also exist in the set of theory-CQs of that argument (that means checking whether they are asking about the same blind spot as any of the CQs generated in Section \ref{sec:theory}). This process leaves us with 4 types of llm-CQs: (i) the ones that are not relevant (CQs 1, 4 and 6 in Figure \ref{fig:methodology}), (ii) the ones that do not match any of the annotated arguments (CQ 3 in Figure \ref{fig:methodology}), (iii) the ones that have a matching argument and a matching theory-CQ (CQ 2 in Figure \ref{fig:methodology}), and (iv) the ones that do have a matching argument but NOT a matching theory-CQ (CQ 5 in Figure \ref{fig:methodology}). We are interested in further investigating this last group, as these are the CQs that the theory did not generate, but are potentially valid.\footnote{In group (ii), there are also potentially valid questions but, since we are not able to relate them to any of the annotated arguments, we do not have a way to validate them. This set can include both invalid questions or valid questions related to non-annotated arguments.} 

Third, the last step to validate this group of LLM-generated CQs consists in assessing their inferential relation to the arguments they have been assigned. That means asking whether it fulfills the core function of CQs: unmasking a blind spot in the argument. We operationalized this evaluation by taking each argument and question pairs and asking: "Can the answer to this question diminish the acceptability of the argument?". The answer to this question can only be \textit{yes} or \textit{no}.\footnote{The guidelines of this evaluation will be published.} In a proof-of-concept evaluation we achieved an IAA of 0.65 with two annotators. 

\section{Results}
\label{sec:results}

In order to generate the critical questions we use two open state-of-the-art LLMs: Llama-2-13B and Zephyr-13B \cite{touvron_llama_2023, tunstall_zephyr_2023}. We employ the instruction-tuned chat versions of the models. For Zephyr, we use the parameters indicated for their chat version and the chat templates used in training. For Llama-2, we use the chat version released in July 2023. With the two prompts, we obtain 495 LLM-generated candidate CQs (llm-CQs).
We now report the results of each of the evaluation steps described in Section \ref{sec:llm-generation}, to later compare the llm-CQs to the theory-CQs, showing the differences between the questions obtained through each of these approaches.

\subsection{Relevance with respect to the Intervention}

The relevance issues found in the llm-CQs are reported in Table \ref{tab:errors}. For all types of issues, Llama-2 works better than Zephyr. While in Llama-2 with the Query prompt 80\% of the CQs are relevant, in Zephyr with the Query+Definition prompt the relevance drops to 30\%.

\begin{table*}[h]
 \footnotesize \centering
\begin{tabular}{lc|ccccc|c}
\hline
\bf Model & \bf Prompt &  \bf Relevant & \bf \bf \begin{tabular}[c]{@{}c@{}}\textbf{New}\\ \textbf{Concept}\end{tabular} & \bf \begin{tabular}[c]{@{}c@{}}\textbf{Bad}\\ \textbf{Reasoning}\end{tabular} &  \bf Non-specific &  \bf Other & \bf TOTAL \\
\hline
Zephyr  &  Q-prompt  & \bf  67.57\% & 14.86\% & 14.86\% & 0.0\% & 2.7\% &  74  \\
Llama-2 &  Q-prompt   &  \bf 80.74\% & 4.44\% & 11.11\% & 2.96\% & 0.74\% &  135  \\
\hline
Zephyr  &  D+Q-prompt   &  \bf 29.46\% & 13.18\% & 7.75\% & 33.33\% & 16.28\% &  129  \\
Llama-2  &  D+Q-prompt   &  \bf 70.7\% & 10.19\% & 6.37\% & 12.1\% & 0.64\% &  157  \\
\hline
All  &  All &  \bf 308 & 50 & 46 & 66 &  25 &  \bf 495 \\
\hline
\end{tabular}
\caption{Relevance issues of the LLM-generated critical questions. By model and prompt. \textit{Q-prompt} refers to the prompt with only the query, and \textit{D+Q-prompt} refers to the prompt that also has the definition of CQs. Each column is one of the relevance issues described in Section \ref{sec:llm-generation}.}
\label{tab:errors}
\end{table*}

When using the prompt with just the query, for both models, over 10\% of the generated questions ask about claims the speaker does not hold (i.e. bad reasoning). Additionally, in Zephyr, 15\% of the questions introduce new concepts. We expected that adding the definition of CQs to the prompt would improve the performance of the models. However, while \textit{bad reasoning} issues are reduced by half for both models, \textit{new concept} issues do not disappear (and even increase for Llama-2). Additionally, a new type of issue is introduced: \textit{non-specific questions}. These are candidate CQs that are not specific to the text, but just general CQs (e.g. "What assumptions is the argument making?"). That is especially the case with Zephyr. With this model, we also get a lot of outputs that are not even questions (the ones classified as \textit{Other}). 

\subsection{Relation of llm-CQs to Arguments and to theory-CQs}

In order to validate the 308 relevant LLM-generated CQs, these need to be related to one of the arguments in the intervention. In this step, the llm-CQs are paired with the arguments of the intervention that prompted them. As a result, 191 unique llm-CQs are associated to at least one of the arguments, resulting in 50 out of the 51 arguments having at least one associated llm-CQ. Since one llm-CQ can be associated with many arguments, and an argument can have multiple associated llm-CQs, the total number of pairs of arguments and llm-CQs is 294. 

Regarding those questions that appeared both in the llm-CQs and in the theory-CQs, we have found 36 unique llm-CQs that have a matching theory-CQ. Since multiple llm-CQs can be associated to one theory-CQ (if they have the same meaning), we obtain 52 pairs of llm-CQs and theory-CQs.

In the end, this step has left us with 242 llm-CQs that are associated to an argument but do not match any of the theory-CQs of that argument.\footnote{Note these are pairs of related llm-CQs and arguments.}  

\subsection{Inferential Validity of the llm-CQs}
\label{sec:validity}

Having related each of the llm-CQs to an argument, we can finally check the validity of each of these critical questions by asking if the answer to the CQ could diminish the acceptability of the argument. We do this with the 242 llm-CQs that have an associated argument but have no matching theory-CQ, since we already know that llm-CQs that matched a theory-CQs are valid critical questions. 

This evaluation results in 64.05\% of the relevant and related llm-CQs being marked as valid (155 questions). The remaining 87 questions do not focus on critical aspects of the argument, often, these ask for additional information that could not impact the acceptability of the argument. 

After the filtering processes described, we have been left with a dataset of 21 interventions associated to three sets of valid CQs: (i) the theory-CQs (129 in total), (ii) the llm-CQs that matched a theory-CQ (52 in total), and (iii) the llm-CQs that did not match a theory-CQ but were found to be valid in Section \ref{sec:validity} (155 in total). That means that we have 207 valid llm-CQs in total (52 plus 155), 137 of which are unique. Therefore, in the end, only 28\% of the 495 candidate llm-CQs end up being relevant and valid (for an example of an intervention in the resulting dataset, see Figure \ref{fig:main}).

\begin{figure*}[h!]
\centering \footnotesize 
\begin{minipage}[b]{1\textwidth}
\centering
\begin{subfigure}[b]{\linewidth}
  \centering 
  \begin{tcolorbox}[boxrule=1pt,colback=white,colframe=mycustomcolor] %
  \textit{MT: "Claire’s absolutely right about that. 
But then the problem is that that form of capitalism wasn’t generating sufficient surpluses. 
And so therefore where did the money flow. 
It didn’t flow into those industrial activities, 
 because in the developed world that wasn’t making enough money."}
  \end{tcolorbox}
  \vspace{-1em}
  \caption{Intervention}
  \label{ex:implicit}
\end{subfigure}
\end{minipage}%
\hfill %

\vspace{0.2cm} %

\begin{minipage}[b]{.49\textwidth}
\begin{subfigure}[b]{\textwidth}
  \centering \scriptsize
  \begin{tcolorbox}[boxrule=1pt,colback=white,colframe=mycustomcolor]
  - How strong is the generalisation that if that form of capitalism was not making enough money in the developed world then the money did not flow into those industrial activities? \\
  - Are there other factors in this particular case that could have interfered with the event of `the money did not flow into those industrial activities'? \\
 - How strong is the generalisation that if that form of capitalism wasn't generating sufficient surpluses then the money did not flow into industrial activities? 
  \end{tcolorbox}
  \vspace{-1em}
  \caption{theory-CQs}
  \label{ex:main}
\end{subfigure}
\end{minipage}
\hfill
\begin{minipage}[b]{.49\textwidth}
\begin{subfigure}[b]{\textwidth}
  \centering \scriptsize
  \begin{tcolorbox}[boxrule=1pt,colback=white,colframe=mycustomcolor]
- How is `sufficient surpluses' defined, and how would one measure it? \\
  - Is MT implying that current forms of capitalism are more successful at generating profits and surpluses than the one being discussed? If yes, why? \\
  - What evidence is there to support the claim that the form of capitalism being used in the developed world was not generating sufficient surpluses? \\
  - Are there any alternative explanations for why the money did not flow into industrial activities? 
\end{tcolorbox}
  \vspace{-1em}
\caption{llm-CQs}
  \label{ex:finance_law_degree}
\end{subfigure}
\end{minipage}
\caption{Example of an instance of the generated reference data. The intervention is from the Moral Maze dataset, and the theory-CQs and the llm-CQs are the result of both of our generation methods. 
}
\label{fig:main}
\end{figure*}

\subsection{Comparing the Approaches}

At this point, it is interesting to study the differences between the sets of questions obtained in each approach. To this goal, all the CQs have been classified regarding the type of blind spot they are trying to unmask. We find that, regarding theory-CQs, the most common type of questions are those asking about the relation between the premises and the conclusion (27\%), followed by questions about the available evidence (24\%), and questions about possible exceptions (18\%). In the case of llm-CQs, asking about evidence is the most common type of CQs (27\%), followed by relations (21\%) and potential consequences of the premises (17\%). Most interestingly, we find that 16\% of llm-CQs are asking for more specific definitions of the concepts present in the argument. This kind of questions are not contemplated at all in the theoretical sets of questions, and both of our annotators considered them valid (the first llm-CQ in Figure \ref{fig:main} is of this type). Finally, the few questions that are generated with both approaches (theory and LLMs) are mostly about consequences and evidence (see Table \ref{tab:types}).

\begin{table}[h!]
\centering \small
\begin{tabular}{r|cl|cl|c}
\hline
\bf Type & \bf t-CQs &  \bf \% & \bf llm-CQs &  \bf \% & \bf match  \\
\hline
evidence & 31 & 24.0 &  55 & 26.6 & 17   \\
relation & 35 & 27.1 &  43 & 20.8 & 10  \\
conseq. & 14 & 10.9 &  35 & 16.9 & 19 \\
definition & 0 & 0.0 & 34 & 16.4 & 0  \\
other & 6 & 4.7 &  20 & 9.7 & 0   \\
alternative & 6 & 4.7 &  7 & 3.4 & 0  \\
exception & 23 & 17.8 & 7 & 3.4 & 5  \\
source & 14 & 10.9 &  6 & 2.9 & 3  \\
\hline
\bf Total & 129 & & 207 & & 52  \\
\hline
\end{tabular}
\caption{Types of questions in the final sets of theory-CQ, valid llm-CQs, and matching CQs between the two approaches. Amount and percentage. The matching ones are also included in the counts of both approaches.}
\label{tab:types}
\end{table}

\section{Concluding Remarks}
\label{sec:discussion}

In this work we have introduced and motivated the task of \textit{Critical Questions Generation}. Moreover, we have studied how to generate valid critical questions with two goals in mind: (i) designing a procedure to obtain reference data, and (ii) discovering the main difficulties that state-of-the-art LLMs face when generating valid critical questions.

Regarding the difficulties of the task, we have found that current LLMs struggle to generate CQs strictly related to the text. On the one hand, they tend to output CQs including new concepts not present in the arguments. On the other hand, they sometimes opt for generating unfiltered lists of very general CQs, with no regard to the given argumentative text. Reasoning is still an issue for these models, and they sometimes struggle to understand what claims are actually held by the given text. 
Finally, while 62\% of the LLM-generated CQs did not have any of these three issues (308 out of 495), only 28\% of the CQs initially generated by LLMs were found to be valid in relation to one of the arguments (137 out of 495), showing that there is a big margin for improvement. 

In relation to the goal of creating a reference dataset, we have shown that the existing theoretical sets of critical questions do not account for all the possible valid critical questions.
In this sense, our results show that only 25\% of the valid llm-CQs had been included in the theoretical sets (52 out of 207). For this reason, we propose 
using both theory-CQs and llm-CQs to build the reference data for this task. Furthermore, we have also observed that the type of questions generated by LLMs differs from the ones created by theory, with the LLMs approach generating many questions related to evidence, consequences and definitions. This suggests that the two approaches (theory and LLMs) are complementary.

While this work has been a first step towards the task of \textit{Critical Questions Generation}, our end goal of automatically generating valid CQs is far from solved. In future work, we will create a larger reference dataset including both theory and llm-CQs to facilitate research on automatic CQs Generation.

Finally, it should be noted that we have not paid any attention to LLM-generated questions that did not match any of the annotated arguments. However, as some arguments might be missing from the annotation (either because they were not in our selected 18 argumentation schemes or because the annotators missed them), some of these questions might be valid CQs. This shows that our work relies heavily on already annotated data with argumentation schemes. And, while the datasets used are reliable \cite{visser_annotating_2021, lawrence_bbc_2018}, there is not a lot of quality data annotated with argumentation schemes, which poses a limitation on how much reference data can be created. As far as we are aware, the only data available is the one detailed in Table \ref{tab:data}, which is all in English.

\section*{Acknowledgements}

This work has been partially supported by the Basque Government (Research group funding IT-1805-22). We are also thankful to the following MCIN/AEI/10.13039/501100011033 projects: (i) DeepKnowledge (PID2021-127777OB-C21) and by FEDER, EU; (ii) Disargue (TED2021-130810B-C21) and European Union NextGenerationEU/PRTR and (iii) DeepMinor (CNS2023-144375) and European Union NextGenerationEU/PRTR.
Blanca Calvo Figueras is supported by the UPV/EHU PIF22/84 predoc grant.

\bibliography{Argument-disinformation}

\clearpage

\appendix

\onecolumn

\pagestyle{empty}

\section{Mapping and Distribution of US2016 and Moral Maze}
\label{annex:distribution}

\begin{table}[!ht]
    \centering \small
    \begin{tabular}{llclc}
    \hline 
        \bf Walton's Argumentation Schemes & \bf Moral Maze Schemes & \bf N & \bf US2016 Schemes & \bf N \\ 
        \hline 
        CauseToEffect & \begin{tabular}[l]{@{}l@{}}CauseToEffect,\\ Cause To Effect\end{tabular}  & 18 & CauseToEffect & 50\\ \arrayrulecolor{gray}\hline
        Consequences & \begin{tabular}[l]{@{}l@{}}NegativeConsequences,\\ PositiveConsequences\end{tabular}  & 15 & Consequences & 38 \\ \arrayrulecolor{gray}\hline
        Example & Example & 11 & Example & 98  \\ \arrayrulecolor{gray}\hline
        Sign & SignFromOtherEvents & 6 & Sign & 40 \\ \arrayrulecolor{gray}\hline
        Analogy & Analogy & 6 & Analogy & 8  \\ \arrayrulecolor{gray}\hline
        PracticalReasoning & PracticalReasoning & 5 & PracticalReasoning & 35  \\ \arrayrulecolor{gray}\hline
        ExpertOpinion & \begin{tabular}[l]{@{}l@{}}ExpertOpinion,\\ Expert Opinion \end{tabular}  & 4 & ExpertOpinion & 4 \\ \arrayrulecolor{gray}\hline
        PopularOpinion & PopularOpinion & 2 & PopularOpinion & 8  \\ \arrayrulecolor{gray}\hline
        CircumstantialAdHominem & CircumstantialAdHominem & 1 & CircumstantialAdHominem & 26  \\ \arrayrulecolor{gray}\hline
        VerbalClassification & ~ & ~ & VerbalClassification & 29  \\ \arrayrulecolor{gray}\hline
        GenericAdHominem & ~ & ~ & GenericAdHominem & 28 \\ \arrayrulecolor{gray}\hline
        PositionToKnow & ~ & ~ & PositionToKnow & 18  \\ \arrayrulecolor{gray}\hline
        Values & ~ & ~ & Values & 14  \\ \arrayrulecolor{gray}\hline
        Bias & ~ & ~ & Bias & 12  \\ \arrayrulecolor{gray}\hline
        FearAppeal & ~ & ~ & FearAppeal & 9  \\ \arrayrulecolor{gray}\hline
        DangerAppeal & ~ & ~ & DangerAppeal & 8 \\ \arrayrulecolor{gray}\hline
        Alternatives & ~ & ~ & Alternatives & 7  \\ \arrayrulecolor{gray}\hline
        PopularPractice & ~ & ~ & PopularPractice & 7  \\ \arrayrulecolor{black}\hline
    \end{tabular}
    \caption{Mapping of labels and distribution of the argumentation schemes in the whole datasets.}
\end{table}

\section{Distribution of the Selected Sample}
\label{annex:distribution_sample}

\begin{table}[h]
\centering \small
\begin{tabular}{lc}
\hline
 \bf Argument scheme &  \bf N  \\
\hline
Argument from Consequences  &  10  \\
Argument from Example  &  6  \\
Practical Reasoning  &  6  \\
Argument from Cause to Effect  &  5  \\
General Ad Hominem  &  5  \\
Circumstantial Ad Hominem  &  5  \\
Argument from Bias  &  4  \\
Argument from Verbal Classification  &  4  \\
Argument from Position to Know  &  3  \\
\hline
 \end{tabular}
  \begin{tabular}{lc}
  \hline
 \bf Argument scheme &  \bf N  \\
\hline
Argument from Sign  &  2  \\
Argument from Values  &  2  \\
Argument from Analogy  &  2  \\
Argument from Expert Opinion  &  1  \\
Argument from  Popular Opinion  &  1  \\
Argument from  Fear Appeal  &  1  \\
Argument from  Alternatives  &  1  \\
Argument from Popular Practice  &  1  \\
Argument from Danger Appeal  &  1  \\
\hline
\end{tabular}
\caption{Distribution of the argumentation schemes in the 21 interventions.}
\label{tab:distribution}
\end{table}

\clearpage

\section{Templates of Critical Questions from Theory}
\label{annex:questions}

\newcolumntype{P}[1]{>{\centering\arraybackslash}m{#1}}

\begin{table}[h!]
\begin{adjustwidth}{-1.5cm}{-1.5cm}
    \centering \footnotesize 
    \begin{tabular}{P{2.2cm}m{15cm}}
    \hline 
        \bf \small Argumentation Scheme & \bf \small Critical Questions \\ 
        \hline 
 Argument From & Is there a proved relation between `$<eventB>$' and `$<eventA>$'? \\
 Sign &  Are there any events other than $<eventB>$ that would more reliably account for $<eventA>$? \\
\hline
 & Is it actually the case that $<subjecta>$ $<featF>$ $<featG>$? Is there evidence on this claim? \\
Argument from Example & Is $<subjecta>$ actually a typical case of other $<subjectx>$ that $<featF>$? How widely applicable is the generalisation? \\
 & Are there special circumstances pertaining to $<subjecta>$ that undermine its generalisability to other $<subjectx>$ that $<featF>$? \\
\hline
Argument  & Is it the case that $<subjecta>$ $<featF>$, or is there room for doubt? \\
 from Verbal    & Is there a proved relation between situations in which "$<featF>$" and situations in which "$<featG>$"? \\
Classification & Is it possible for the particular case of $<subjecta>$ that $<featG>$ is not the case? \\
\hline

 Argument & Is $<subjecta>$ in a position to know whether $<eventA>$? \\
 from Position  & Is $<subjecta>$ an honest (trustworthy, reliable) source? \\
to Know & Did $<subjecta>$ assert that $<eventA>$? \\
\hline
& Is $<expertE>$ a genuine expert in $<domainD>$? \\
& Did $<expertE>$ really assert that $<eventA>$? \\
 & Is $<expertE>$'s pronouncement directly quoted? If not, is a reference to the original source given? Can it be checked? \\
Argument from    & If $<expertE>$'s advice is not quoted, does it look like important information or qualifications may have been left out? \\
Expert Opinion & Is what $<expertE>$ said clear? Are there technical terms used that are not explained clearly? \\
 & Is $<eventA>$ relevant to domain $<domainD>$? \\
 & Is $<eventA>$ consistent with what other experts in $<domainD>$ say? \\
 & Is $<eventA>$ consistent with known evidence in $<domainD>$? \\
\hline

Argument from & How strong is the generalisation that if $<eventA>$ then $<eventB>$? \\
 Cause to Effect & Are there other factors in this particular case that could have interfered with the event of `$<eventB>$'? \\
\hline
Argument from    & If $<eventA>$, will $<eventB>$ occur? What evidence supports this claim? How likely are the consequences? \\
Consequences & What other consequences should also be taken into account if $<neg>$ $<eventA>$? \\
\hline
 & Are $<C1>$ and $<C2>$ similar in the respect cited? \\
Argument from  & Is $<eventA>$ true in $<C1>$? \\
Analogy  & Are there differences between $<C1>$ and $<C2>$ that would tend to undermine the force of the similarity cited? \\
 & Is there some other case that is also similar to $<C1>$, but in which $<eventA>$ is false? \\

\hline
Argument from & What evidence supports that $<eventA>$ is generally accepted as true? \\
 Popular Opinion & Even if $<eventA>$ is generally accepted as true, are there any good reasons for doubting that it is true? \\
\hline
 Argument from & What actions or other indications show that $<large\_majority>$ accept that $<eventA>$ is the right thing to do? \\
 Popular Practice & Even if $<large\_majority>$ accepts $<eventA>$ is the right thing to do, are there grounds for thinking they are justified in accepting it as a prudent course of action? \\

\hline
Argument  from Bias  & What evidence is there that $<subjecta>$ is $<subjectx>$? Could $<subjecta>$ have taken evidence on many sides even if $<subjecta>$ is $<subjectx>$? \\
 & Does the matter of $<eventA>$ require $<subjecta>$ to take evidence on many sides? \\

\hline
Generic Ad & How does the allegation made affect the reliability of $<subjecta>$? \\
 Hominem & Is the reliability of $<subjecta>$ relevant in the current dialogue? \\

\hline
Practical & Are there other relevant goals that conflict with $<goalG>$? \\
 Reasoning & Are there alternative actions to $<eventA>$ to achieve $<goalG>$? If so, which is the most efficient action? \\
 & Could $<eventA>$ have consequences that we should take into account? Is it practically possible? \\

\hline
Argument from & Can $<eventB>$ happen even if $<eventA>$ is the case? \\
 Alternatives & Is $<eventA>$ plausibly not the case? What evidence supports this claim? \\
\hline
 & If $<eventA>$, will $<eventB>$ occur? What evidence supports this claim? \\
 Argument from & Why is $<eventB>$ a danger? To whom is $<eventB>$ a danger? \\
Danger Appeal  & Is there a way of preventing $<eventA>$? \\
 & Are there other consequences of preventing $<eventA>$ that we should take into account? \\

\hline
& Is $<valueV>$ seen as $<direction>$ for most people? \\
 Argument from  & Are there reasons to believe that $<valueV>$ is not $<direction>$ in this situation? \\
 Values  & Will a subject that sees $<valueV>$ as not $<direction>$ agree with retaining commitment to "$<goalG>$"? \\
\hline
& Is $<eventB>$ bad? Why and to whom is it bad? \\
 Argument from  & Is $<eventA>$ a way to prevent $<eventB>$? \\
 Fear Appeal & Is it practically possible for $<eventA>$ to happen? \\
 & Are there other consequences from $<eventA>$? \\
\hline
 & Does $<argument1>$ imply $<eventA>$? \\
Circumstantial Ad Hominem & Can the practical inconsistency between $<subjecta>$'s commitments and "$<eventA>$" be identified? Can it be shown by evidence? Could it be explained by further dialogue? \\
 & Does the inconsistency between $<subjecta>$'s commitments and "$<eventA>$" result in a decrease of credibility for $<subjecta>$? Does $<subjecta>$'s argument depend in its credibility in this context? \\
\hline
    \end{tabular}
    \caption{Sets of templates of critical question for each argumentation scheme in \citet{walton_argumentation_2008}. Some have been reformulated to sound more natural. }
\end{adjustwidth}
\end{table}

\clearpage

\end{document}


\usepackage{caption}
\usepackage{graphicx}
\usepackage{float}
\usepackage{svg}
\usepackage{amsmath} 
\usepackage{lipsum}
\usepackage{colortbl}
\usepackage{booktabs}
\usepackage{nicematrix}
\usepackage[listings]{tcolorbox}

\definecolor{mycustomcolor}{HTML}{1F77B4}

\appendix

\onecolumn

\section{Mapping and Distribution of US2016 and Moral Maze}
\label{annex:distribution}

\begin{table}[!ht]
    \centering \small
    \begin{tabular}{llclc}
    \hline 
        \bf Argumentation Schemes & \bf Moral Maze Schemes & \bf N & \bf US2016 Schemes & \bf N \\ 
        \hline 
        CauseToEffect & \begin{tabular}[l]{@{}l@{}}CauseToEffect,\\ Cause To Effect\end{tabular}  & 18 & CauseToEffect & 50\\ \arrayrulecolor{gray}\hline
        Consequences & \begin{tabular}[l]{@{}l@{}}NegativeConsequences,\\ PositiveConsequences\end{tabular}  & 15 & Consequences & 38 \\ \arrayrulecolor{gray}\hline
        Example & Example & 11 & Example & 98  \\ \arrayrulecolor{gray}\hline
        Sign & SignFromOtherEvents & 6 & Sign & 40 \\ \arrayrulecolor{gray}\hline
        Analogy & Analogy & 6 & Analogy & 8  \\ \arrayrulecolor{gray}\hline
        PracticalReasoning & PracticalReasoning & 5 & PracticalReasoning & 35  \\ \arrayrulecolor{gray}\hline
        ExpertOpinion & \begin{tabular}[l]{@{}l@{}}ExpertOpinion,\\ Expert Opinion \end{tabular}  & 4 & ExpertOpinion & 4 \\ \arrayrulecolor{gray}\hline
        PopularOpinion & PopularOpinion & 2 & PopularOpinion & 8  \\ \arrayrulecolor{gray}\hline
        CircumstantialAdHominem & CircumstantialAdHominem & 1 & CircumstantialAdHominem & 26  \\ \arrayrulecolor{gray}\hline
        VerbalClassification & ~ & ~ & VerbalClassification & 29  \\ \arrayrulecolor{gray}\hline
        GenericAdHominem & ~ & ~ & GenericAdHominem & 28 \\ \arrayrulecolor{gray}\hline
        PositionToKnow & ~ & ~ & PositionToKnow & 18  \\ \arrayrulecolor{gray}\hline
        Values & ~ & ~ & Values & 14  \\ \arrayrulecolor{gray}\hline
        Bias & ~ & ~ & Bias & 12  \\ \arrayrulecolor{gray}\hline
        FearAppeal & ~ & ~ & FearAppeal & 9  \\ \arrayrulecolor{gray}\hline
        DangerAppeal & ~ & ~ & DangerAppeal & 8 \\ \arrayrulecolor{gray}\hline
        Alternatives & ~ & ~ & Alternatives & 7  \\ \arrayrulecolor{gray}\hline
        PopularPractice & ~ & ~ & PopularPractice & 7  \\ \arrayrulecolor{black}\hline
    \end{tabular}
    \caption{Mapping of labels and distribution of the argumentation schemes in the whole datasets.}
\end{table}

\section{Distribution of the Selected Sample}
\label{annex:distribution_sample}

\begin{table}[h]
\centering \small
\begin{tabular}{lc}
\hline
 \bf Argument scheme &  \bf N  \\
\hline
Argument from Consequences  &  10  \\
Argument from Example  &  6  \\
Practical Reasoning  &  6  \\
Argument from Cause to Effect  &  5  \\
General Ad Hominem  &  5  \\
Circumstantial Ad Hominem  &  5  \\
Argument from Bias  &  4  \\
Argument from Verbal Classification  &  4  \\
Argument from Position to Know  &  3  \\
\hline
 \end{tabular}
  \begin{tabular}{lc}
  \hline
 \bf Argument scheme &  \bf N  \\
\hline
Argument from Sign  &  2  \\
Argument from Values  &  2  \\
Argument from Analogy  &  2  \\
Argument from Expert Opinion  &  1  \\
Argument from  Popular Opinion  &  1  \\
Argument from  Fear Appeal  &  1  \\
Argument from  Alternatives  &  1  \\
Argument from Popular Practice  &  1  \\
Argument from Danger Appeal  &  1  \\
\hline
\end{tabular}
\caption{Distribution of the argumentation schemes in the 21 interventions.}
\label{tab:distribution}
\end{table}

\section{Types of Critical Questions}
\label{annex:types}

\begin{table*}[h!]
\centering \small
\begin{tabular}{r|cc|cc|c}
\hline
\bf Type & \bf Nº in theory-CQs &  \bf \% & \bf Nº in llm-CQs &  \bf \% & \bf Nº matching  \\
\hline
evidence & 31 & 24.03 &  55 & 26.57 & 17   \\
relation & 35 & 27.13 &  43 & 20.77 & 10  \\
consequences & 14 & 10.85 &  35 & 16.91 & 19 \\
definition & 0 & 0.0 & 34 & 16.43 & 0  \\
other & 6 & 4.65 &  20 & 9.66 & 0   \\
alternative & 6 & 4.65 &  7 & 3.38 & 0  \\
exception & 23 & 17.83 & 7 & 3.38 & 5  \\
source & 14 & 10.85 &  6 & 2.9 & 3  \\
\hline
\bf Total & 129 & & 207 & & 52  \\
\hline
\end{tabular}
\caption{Types of questions in the final sets of theory-CQ, valid llm-CQs, and matching CQs between the two approaches. Amount and percentage. The matching ones are also included in the counts of both approaches.}
\label{tab:types}
\end{table*}

\clearpage

\section{Prompts}
\label{annex:prompts}




\begin{figure}[h]
\centering
\footnotesize
\begin{minipage}[b]{0.49\columnwidth}
\centering
\begin{subfigure}[b]{\columnwidth}
  \centering
  \begin{tcolorbox}[boxrule=1pt,colback=white,colframe=mycustomcolor] %
  List the critical questions that should be asked regarding the arguments in the following paragraph: \\ \textit{{SPEAKER}: “{INTERVENTION}”}
  \end{tcolorbox}
  \caption{Q-prompt}
  \label{ex:implicit}
\end{subfigure}
\end{minipage}%
\hfill %
\begin{minipage}[b]{0.49\columnwidth}
\centering
\begin{subfigure}[b]{\columnwidth}
  \centering
  \begin{tcolorbox}[boxrule=1pt,colback=white,colframe=mycustomcolor]
  Critical questions are the set of enquiries that should be asked in order to judge if an argument is good or fallacious by unmasking the assumptions held by the premises of the argument. List the critical questions that should be asked regarding the arguments in the following paragraph: \\ \textit{{SPEAKER}: “{INTERVENTION}”}
  \end{tcolorbox}

  \caption{D+Q-prompt}
  \label{ex:political}
\end{subfigure}
\end{minipage}

\caption{Prompts used for the experiments. The first prompt includes the query (Q-prompt), the second prompt includes the query and the definition of CQs (D+Q-prompt).}
\label{fig:main}
\end{figure}

\section{Example of the Generated Reference Dataset}
\label{annex:dataset}

\begin{table*}[htbp]
\centering \small
\begin{NiceTabular}{p{5.5cm}p{10cm}}[]
\cline{1-2}
 \bf Intervention &    \\ 
 \hline
\Block{10-1}{
MT: "Claire’s absolutely right about that. 
But then the problem is that that form of capitalism wasn’t generating sufficient surpluses. 
And so therefore where did the money flow. 
It didn’t flow into those industrial activities, 
 because in the developed world that wasn’t making enough money."
} & \bf Theory CQs \\
\cline{2-2}
 & - How strong is the generalisation that if that form of capitalism was not making enough money in the developed world then the money did not flow into those industrial activities? \\
 & - Are there other factors in this particular case that could have interfered with the event of `the money did not flow into those industrial activities'? \\
 & - How strong is the generalisation that if that form of capitalism wasn't generating sufficient surpluses then the money did not flow into industrial activities? \\
 & - Are there other factors in this particular case that could have interfered with the event of `the money did not flow into industrial activities'? \\
\cline{2-2}
& \bf LLM CQs   \\ 
\cline{2-2}
 & - How is `sufficient surpluses' defined, and how would one measure it? \\
 & - Is MT implying that current forms of capitalism are more successful at generating profits and surpluses than the one being discussed? If yes, why? \\
 & - What evidence is there to support the claim that the form of capitalism being used in the developed world was not generating sufficient surpluses? \\
 & - Are there any alternative explanations for why the money did not flow into industrial activities? \\
 & - What alternative explanations are there for the lack of investment in industrial activities? \\
\hline
\end{NiceTabular}
\caption{Example of an instance of the generated reference data. The intervention is from the Moral Maze dataset, and the theory-CQs and the llm-CQs are the result of both of our generation methods. }
\label{table2}
\end{table*}

\clearpage

\section{Templates of Critical Questions from Theory}
\label{annex:questions}

\newcolumntype{P}[1]{>{\centering\arraybackslash}m{#1}}

\begin{table*}[h!]
    \centering \scriptsize 
    \begin{tabular}{P{2cm}m{12cm}}
    \hline 
        \bf \small Argumentation Scheme & \bf \small Critical Questions \\ 
        \hline 
Argument From & Is there a proved relation between `$<eventB>$' and `$<eventA>$'? \\
 Sign &  Are there any events other than $<eventB>$ that would more reliably account for $<eventA>$? \\
\hline
 & Is it actually the case that $<subjecta>$ $<featF>$ $<featG>$? Is there evidence on this claim? \\
Argument from Example & Is $<subjecta>$ actually a typical case of other $<subjectx>$ that $<featF>$? How widely applicable is the generalisation? \\
 & Are there special circumstances pertaining to $<subjecta>$ that undermine its generalisability to other $<subjectx>$ that $<featF>$? \\
\hline
Argument  & Is it the case that $<subjecta>$ $<featF>$, or is there room for doubt? \\
 from Verbal    & Is there a proved relation between situations in which "$<featF>$" and situations in which "$<featG>$"? \\
Classification & Is it possible for the particular case of $<subjecta>$ that $<featG>$ is not the case? \\
\hline

 Argument & Is $<subjecta>$ in a position to know whether $<eventA>$? \\
 from Position  & Is $<subjecta>$ an honest (trustworthy, reliable) source? \\
to Know & Did $<subjecta>$ assert that $<eventA>$? \\
\hline
& Is $<expertE>$ a genuine expert in $<domainD>$? \\
& Did $<expertE>$ really assert that $<eventA>$? \\
 & Is $<expertE>$'s pronouncement directly quoted? If not, is a reference to the original source given? Can it be checked? \\
Argument from    & If $<expertE>$'s advice is not quoted, does it look like important information or qualifications may have been left out? \\
Expert Opinion & Is what $<expertE>$ said clear? Are there technical terms used that are not explained clearly? \\
 & Is $<eventA>$ relevant to domain $<domainD>$? \\
 & Is $<eventA>$ consistent with what other experts in $<domainD>$ say? \\
 & Is $<eventA>$ consistent with known evidence in $<domainD>$? \\
\hline

Argument from & How strong is the generalisation that if $<eventA>$ then $<eventB>$? \\
 Cause to Effect & Are there other factors in this particular case that could have interfered with the event of `$<eventB>$'? \\
\hline
Argument from    & If $<eventA>$, will $<eventB>$ occur? What evidence supports this claim? How likely are the consequences? \\
Consequences & What other consequences should also be taken into account if $<neg>$ $<eventA>$? \\
\hline
 & Are $<C1>$ and $<C2>$ similar in the respect cited? \\
Argument from  & Is $<eventA>$ true in $<C1>$? \\
Analogy  & Are there differences between $<C1>$ and $<C2>$ that would tend to undermine the force of the similarity cited? \\
 & Is there some other case that is also similar to $<C1>$, but in which $<eventA>$ is false? \\

\hline
Argument from & What evidence supports that $<eventA>$ is generally accepted as true? \\
 Popular Opinion & Even if $<eventA>$ is generally accepted as true, are there any good reasons for doubting that it is true? \\
\hline
 Argument from & What actions or other indications show that $<large\_majority>$ accept that $<eventA>$ is the right thing to do? \\
 Popular Practice & Even if $<large\_majority>$ accepts $<eventA>$ is the right thing to do, are there grounds for thinking they are justified in accepting it as a prudent course of action? \\


\hline
Argument  from Bias  & What evidence is there that $<subjecta>$ is $<subjectx>$? Could $<subjecta>$ have taken evidence on many sides even if $<subjecta>$ is $<subjectx>$? \\
 & Does the matter of $<eventA>$ require $<subjecta>$ to take evidence on many sides? \\

\hline
Generic Ad & How does the allegation made affect the reliability of $<subjecta>$? \\
 Hominem & Is the reliability of $<subjecta>$ relevant in the current dialogue? \\

\hline
& Are there other relevant goals that conflict with $<goalG>$? \\
Practical Reasoning & Are there alternative actions to $<eventA>$ to achieve $<goalG>$? If so, which is the most efficient action? \\
 & Could $<eventA>$ have consequences that we should take into account? Is it practically possible? \\

\hline
Argument from & Can $<eventB>$ happen even if $<eventA>$ is the case? \\
 Alternatives & Is $<eventA>$ plausibly not the case? What evidence supports this claim? \\
\hline
 & If $<eventA>$, will $<eventB>$ occur? What evidence supports this claim? \\
 Argument from & Why is $<eventB>$ a danger? To whom is $<eventB>$ a danger? \\
Danger Appeal  & Is there a way of preventing $<eventA>$? \\
 & Are there other consequences of preventing $<eventA>$ that we should take into account? \\

\hline
& Is $<valueV>$ seen as $<direction>$ for most people? \\
 Argument from  & Are there reasons to believe that $<valueV>$ is not $<direction>$ in this situation? \\
 Values  & Will a subject that sees $<valueV>$ as not $<direction>$ agree with retaining commitment to "$<goalG>$"? \\
\hline
& Is $<eventB>$ bad? Why and to whom is it bad? \\
 Argument from  & Is $<eventA>$ a way to prevent $<eventB>$? \\
 Fear Appeal & Is it practically possible for $<eventA>$ to happen? \\
 & Are there other consequences from $<eventA>$? \\
\hline
 & Does $<argument1>$ imply $<eventA>$? \\
Circumstantial Ad Hominem & Can the practical inconsistency between $<subjecta>$'s commitments and "$<eventA>$" be identified? Can it be shown by evidence? Could it be explained by further dialogue? \\
 & Does the inconsistency between $<subjecta>$'s commitments and "$<eventA>$" result in a decrease of credibility for $<subjecta>$? Does $<subjecta>$'s argument depend in its credibility in this context? \\
\hline
    \end{tabular}
    \caption{Sets of templates of critical question for each argumentation scheme in \citet{walton_argumentation_2008}. Some have been reformulated to sound more natural. }
\end{table*}

\clearpage